\documentclass[10pt,twocolumn,letterpaper,pagebackref,breaklinks,colorlinks,allcolors=cvprblue]{article}

\usepackage{cvpr}              
\usepackage[pagebackref,breaklinks,colorlinks,allcolors=cvprblue]{hyperref}
\usepackage{graphicx}
\usepackage{booktabs}
\usepackage{orcidlink}
\usepackage{multirow}
\usepackage{threeparttable}
\usepackage{adjustbox}
\usepackage{wrapfig}

\definecolor{cvprblue}{rgb}{0.21,0.49,0.74}

\def\paperID{*****} 
\def\confName{arXiv}
\def\confYear{2026}

\title{MatchingPolicy: Correspondence-Aware Policy Enables Cross-Object In-Context Learning}
\vspace{-1mm}
\author{
\begin{tabular}{@{}cccc@{}}
Qijin She\footnotemark[1]
&
Hanyang Yu\footnotemark[1]
&
Zeming Li
&
Ping Tan\footnotemark[2]
\\
{\ttfamily\scriptsize qijinshe@outlook.com}
&
{\ttfamily\scriptsize hanyangyu1021@gmail.com}
&
{\ttfamily\scriptsize zlikp@connect.ust.hk}
&
{\ttfamily\scriptsize pingtan@ust.hk}
\\[0.6em]
\multicolumn{4}{c}{
Hong Kong University of Science and Technology
}
\\
\multicolumn{4}{c}{
Hong Kong SAR, China
}
\end{tabular}
}

\begin{document}
\maketitle
\begingroup
\renewcommand{\thefootnote}{\fnsymbol{footnote}}
\footnotetext[1]{Equal contribution.}
\footnotetext[2]{Corresponding author.}
\endgroup

\begin{abstract}
  In-context imitation learning enables few-shot policy generalization but struggles to maintain performance on unseen objects and novel scenarios. To address this, we introduce MatchingPolicy, a correspondence-driven framework that explicitly decouples demonstration-to-scene matching from policy learning. Central to our method is a correspondence-aware diffusion policy that conditions robotic actions directly on dense semantic correspondences. This architectural separation resolves the inherent conflict between correspondence identification and action adaptation, enabling robust out-of-distribution transfer. Our framework integrates vision foundation models with a novel two-stage matching algorithm to dynamically establish reliable correspondences. Extensive evaluations on RLBench and real-world manipulation tasks confirm that MatchingPolicy achieves superior few-shot performance, generalizing reliably across unseen object instances and semantic categories. Visualization results are available at \url{https://matchingpolicy.github.io}.
\end{abstract}

\vspace{-1mm}
\section{Introduction}

Imitation learning has demonstrated remarkable potential in acquiring diverse manipulation skills. However, traditional approaches typically require a large amount of costly, task-specific demonstration data to train a policy for even a single skill. While recent efforts have sought to train universal, multi-task imitation policies ~\cite{kim2024openvla} on large and diverse robotic datasets~\cite{o2024open}, these models still struggle to generalize to entirely new tasks, and often depend on task-specific fine-tuning to reach acceptable performance.
Inspired by breakthroughs in language and vision, in-context learning has recently been introduced to robotic policy learning~\cite{fu2024context} as a way to bypass fine-tuning. In this paradigm, a model is conditioned at test time on a few task demonstrations that provide fine-grained, step-by-step guidance, rather than coarse natural language commands, enabling the policy to adapt to unseen tasks on the fly.

Despite this promise, prior in-context learning approaches for robotics often fail to generalize to novel objects and tasks, even when the demonstrations feature the exact unseen task and object~\cite{jain2024vid2robot}. We argue that a key reason for this shortcoming is that current methods ask a single model to simultaneously: (1) infer precise correspondences between the demonstration and the current scene from scratch using limited robotic-domain data, and (2) adapt the demonstrated motions to the new objects and scenes. This dual burden can hinder both correspondence extraction and action adaptation, especially in visually diverse or geometrically complex environments.

To address this challenge, we propose MatchingPolicy, a novel in-context imitation learning framework for robotic manipulation that explicitly decouples correspondence extraction from policy learning. Rather than forcing the policy to implicitly discover correspondences, we leverage explicitly pre-computed correspondences and focus on training a policy that adapts robot behavior accordingly. Additionally, we introduce an online adaptive semantic matching algorithm that dynamically establishes dense and reliable correspondences between demonstrations and the current scene.

Our correspondence-aware policy offers three key advantages. First, by leveraging explicit correspondences between the demonstration and the current scene, the policy can more precisely localize task-relevant interaction regions, enabling more accurate and adaptive action generation.
Second, as the policy operates on correspondence features rather than raw visual object representations, it becomes inherently object-agnostic, substantially enhancing its ability to generalize across different object instances and categories. Third, the decoupled design, separating correspondence extraction from policy learning, facilitates the seamless integration of powerful off-the-shelf visual and semantic matching methods~\cite{oquab2023dinov2, zhang2023tale} without retraining.

Our results show that the proposed MatchingPolicy outperforms the state-of-the-art in-context imitation learning method on the RLBench~\cite{james2020rlbench}. We further validate our approach on a set of challenging real-world manipulation tasks, where it demonstrates strong generalization to a wide variety of objects. The main contributions are as follows:
\begin{itemize}
\item We propose a novel approach that separates cross-scene correspondence extraction from policy learning, empowered by a correspondence-aware policy conditioned on the extracted explicit correspondences, thereby
enhancing flexibility and scalability.
\item We design a novel diffusion policy biased to leverage direct relationships from explicit correspondences, improving the model’s generalization across tasks and objects.
\item We introduce an online adaptive matching algorithm that dynamically provides reliable correspondences during execution, enabling precise real-world policy adaptation. 
\item We conduct extensive evaluations of our method in both simulation and real-world settings. The results demonstrate that the proposed approach achieves strong few-shot performance, while also exhibiting robust cross-instance and even cross-category generalization.
\end{itemize}    
\section{Related Work}

\subsection{Multi-task Imitation Learning}
A growing body of work focuses on developing generalist robot policies~\cite{kim2024openvla, team2024octo, jang2022bc, black2410pi0, liu2024rdt, cheang2024gr, lin2026universal}. These models, typically conditioned on visual observations and language instructions, are trained on large-scale robotic datasets in real world ~\cite{o2024open, khazatsky2024droid, bu2025agibot} or in simulation ~\cite{james2020rlbench, mu2024robotwin} and achieve strong performance on the tasks seen during training. With appropriate visual inputs and language prompts, they can generalize zero-shot to related tasks and scenes under varying visual conditions. However, their performance often degrades on entirely unseen tasks, additional fine-tuning is usually required to attain acceptable results. Our method bypasses the task generalization obstacle by directly conditioning the policy on demonstration, which showcases the fine-grained intention of the task more effectively.

\subsection{In-context Imitation Learning} In-context learning (ICL) has emerged as a powerful paradigm, enabling models to adapt to novel tasks of language  ~\cite{brown2020language}, vision ~\cite{zhou2024visual} and sequential decision making ~\cite{raparthy2023generalization} from only a handful of demonstrations without retraining. In robotics, explicit ICL-based approaches often begin by establishing correspondences between given demonstrations and a new scene, followed by transforming the trajectories using handcrafted heuristics~\cite{zhang2024one, tang2025functo, zhu2024vision, heppert2024ditto}. 
While effective in controlled settings, they typically demand considerable human effort 
and their adaptability to truly novel scenarios remains constrained. The in-context capability of LLMs has also been adapted to robotic action prediction ~\cite{di2024keypoint, yin2024context}. However, these methods usually run in open-loop and face difficulty to efficiently predict accurate motions.

Recent studies have pursued training in-context policy models directly, leveraging transformers~\cite{jain2024vid2robot, fu2024context} or graph neural networks~\cite{vosylius2024instant} to minimize human involvement. 
In contrast, MatchingPolicy adopts a direct and modular strategy: it explicitly extracts correspondence information from Vision Foundation Models (VFMs), freeing the policy to focus solely on adapting robot actions to the new environment. This decoupled design not only reduces manual intervention but also enhances generalization to novel objects.

Recent works \cite{sridhar2025ricl, sridhar2024regent} employ a similar modular design, decoupling relationship extraction between demonstrations and the current scene from action prediction using retrieval-augmented generation. In contrast, our method extracts dense point-wise correspondences, which not only provide a similarity measure but also delivers fine-grained spatial cues that facilitate robot action prediction.

\subsection{Keypoint-based Policy Learning}
Keypoint-based representations are widely used in robotics because of the sparsity and interpretability~\cite{puang2020kovis, gao2023k}. They also serve as high-level abstract action representations, either predicted by large language models (LLMs)~\cite{huang2024rekep, huang2024copa} or generated by learned policies~\cite{wen2023anypoint, bharadhwaj2024track2act, xu2024flow}, and can be mapped to low-level robotic control, enabling LLM-driven manipulation and cross-embodiment policy transfer.

Recently, several works have integrated keypoint-based representation into policy training as observations ~\cite{haldar2025point, wang2024gendp, levy2024p3, wang2025skil, fang2025kalm}, demonstrating improved generalization to novel scenes and objects and improving the data efficiency.
These methods typically rely on pre-defined keypoints, either selected manually or determined via unsupervised clustering, which can limit the policy's adaptability to novel scenarios.
Our work differs in two key aspects: 1) Task Scope: prior works focus on single-task imitation learning, whereas MatchingPolicy targets in-context multi-task learning; 2) Key Point Adaptation: instead of static pre-defined key points, our method dynamically generates key points based on matching confidence and visibility. This eliminates human effort while ensuring better feature alignment.

\section{Method}

\begin{figure*}[t!]
    \centering
    \includegraphics[width=0.92\textwidth]{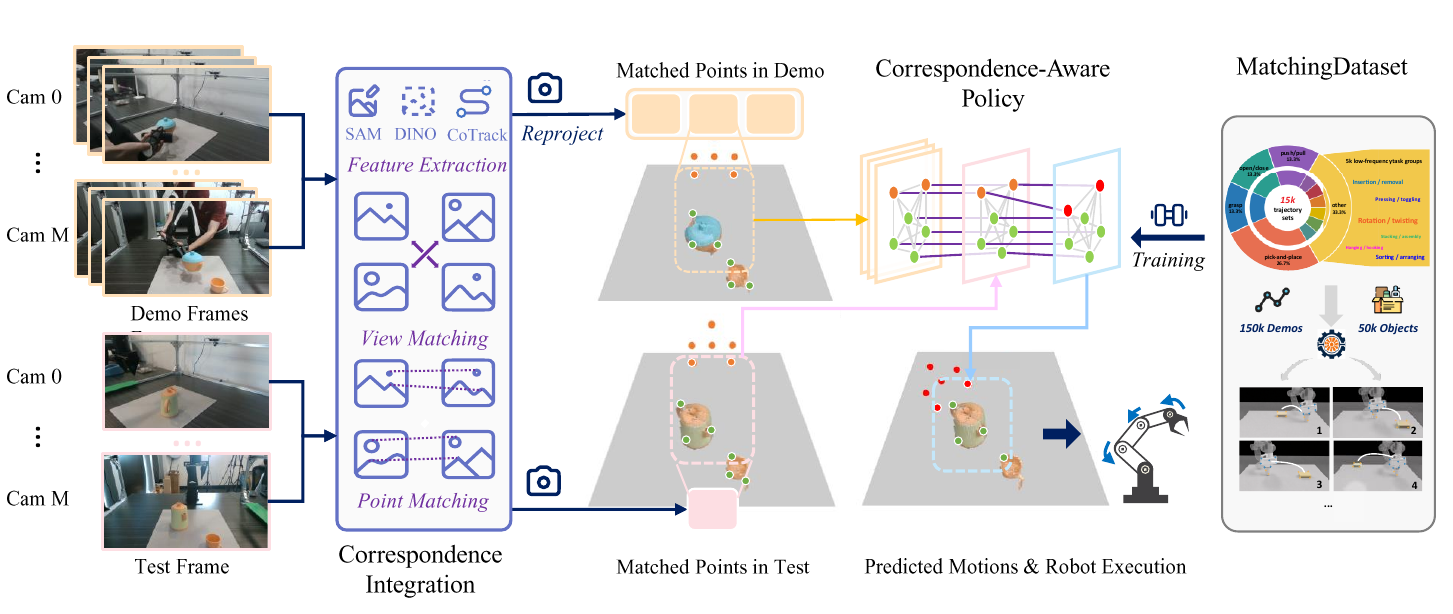} 
    \caption{The overview of MatchingPolicy. Correspondence points are extracted using a two-stage feature matching algorithm with VFMs and subsequently passed into a graph-based diffusion policy model to predict 6D gripper poses. These predicted motions are then translated into robot joint commands, with the model trained end-to-end on our synthetic dataset.}
    \label{fig:pipe}\vspace{-1mm}
\end{figure*}

\paragraph{Problem Formulation}
We address robotic manipulation in a few-shot, in-context learning setting. Given a small set of $N$ demonstrations $\{D_j\}_{j=1}^N$, where $N \in \{1,2\}$, the goal is to leverage these demonstrations to perform the same task depicted in the demos, but in a novel scene.

The policy observation of the current scene is represented by a state $s_c = (P_c, g_c)$. Here, $P_c$ denotes the 3D point cloud, obtained from RGB-D images and transformed into the gripper’s local coordinate frame via the gripper pose $T_c \in \mathrm{SE}(3)$, while $g_c \in \{0, 1\}$ represents a binary gripper state.
Similarly, a demonstration $D_j$ is a time-indexed sequence of $L$ states, $D_j = (s_{j,t})_{t=1}^L$, where each $s_{j,t} = (P_{j,t}, g_{j,t})$ follows the same definition as $s_c$. Given the current state $s_c$ and the demonstration set $\{D_j\}_{j=1}^N$, the policy $\pi$ predicts a horizon of $K$ future actions $(a_k)_{k=1}^K$, where $a_k = (\Delta T_k, g_k)$ comprises a relative gripper transformation $\Delta T_k$ and a target gripper state $g_k$.  Formally, the objective is to learn a policy that models the conditional distribution $\pi = p(a_{1:K} \mid s_c, (D_j)_{j=1}^N)$. For convenience, we use the term "frame" to refer uniformly to the current state, a demonstration state, or a predicted action, drawing a direct analogy to frames in a video.

\paragraph{Method Overview}

The pipeline of our method is shown in \cref{fig:pipe}. 
Our key insight is to decouple correspondence extraction from action prediction. 
Rather than requiring the policy to infer demonstration-current alignment implicitly, we first explicitly establish semantic correspondences between the current observation and the demonstration frames. 
Based on these correspondences, we select a sparse set of keypoints in the current frame together with their matched counterparts across the demonstrations. 
These correspondence-indexed point sets provide a structured context that captures task-relevant spatial and temporal cues from the demonstrations.
Given this structured context, our correspondence-aware policy predicts future end-effector motions for the current scene. 
The policy treats the established correspondences as an explicit prior for routing information from demonstrations to the current observation and to future action predictions, enabling it to generate actions consistent with the demonstrated intent. 
The predicted end-effector motions are then converted into executable robot actions with inverse kinematics.

During training, we use our synthetic demonstration dataset with ground-truth correspondences (\cref{sec:matching_dataset}) to supervise the correspondence-aware policy (\cref{sec:graph}). 
At deployment time, where ground-truth correspondences are unavailable, we combine vision foundation models with visual point trackers to extract semantic and spatio-temporal correspondences from real-world observations (\cref{sec:real}). 
The resulting correspondences are used by the same policy without changing the action prediction module.

\subsection{Correspondence-Aware Diffusion Policy } 
\begin{figure*}[t!]
    \centering
    \includegraphics[width=0.92\textwidth]{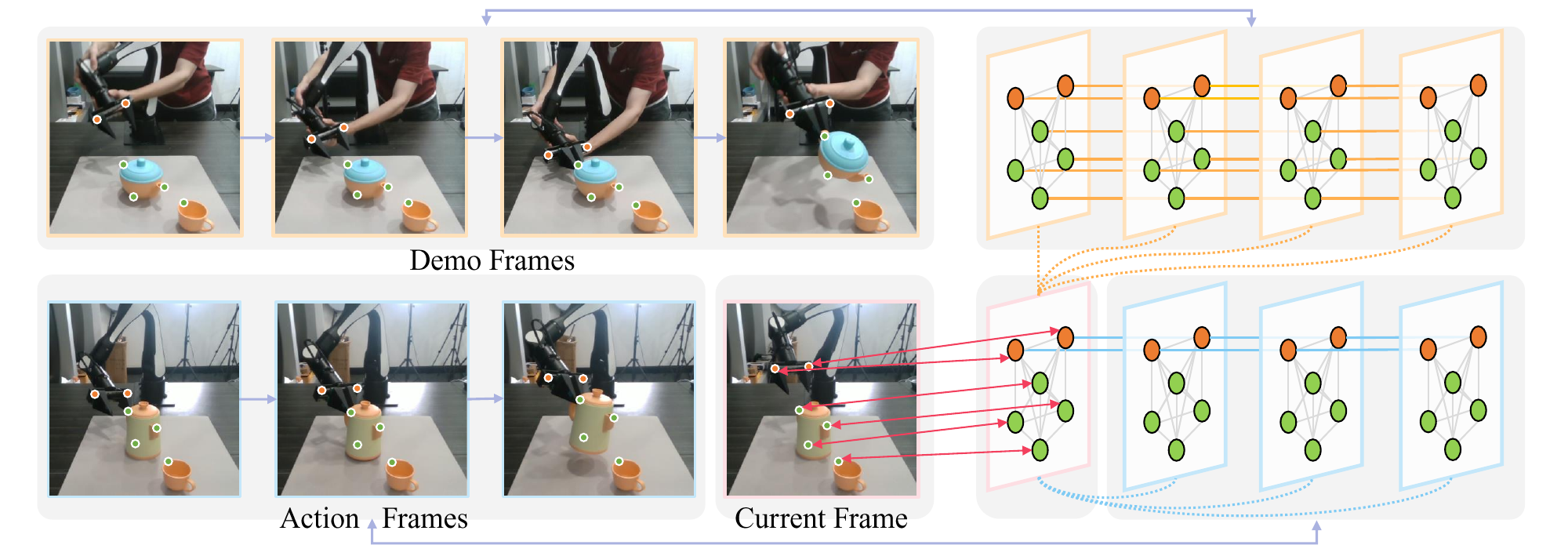} 
    \caption{Our Correspondence-Aware Policy Model. Left: The selected correspondence key points on 2D images of demonstration frames, current frame, and action frames. Right: The structure of our graph-based policy model. The red lines with two-way arrow indicate the one-to-one mapping between the graph nodes in the network and the key points in the current frame. Node lines indicate data transfer. The lines connecting the two frames are a simplified view; the detailed node-to-node links are omitted for visual clarity}\vspace{-1mm}
    \label{fig:graph}
\end{figure*}
\label{sec:graph}

\paragraph{Keypoint Selection with Correspondence.}
To reduce network burden and improve robustness,
we select scene keypoints from the current frame by considering both geometric coverage and correspondence reliability across demonstrations. 
Specifically, a point is retained as a candidate only if it has semantically matched counterparts in most demonstration frames, which filters out unstable matches caused by occlusions or errors from vision foundation models. 
We then apply Farthest Point Sampling (FPS) to the candidate set to obtain a spatially representative and diverse subset of scene keypoints, and select their matched points in the demonstration frames accordingly.

In addition to scene keypoints, we include a fixed set of gripper keypoints to represent the gripper pose and state. 
Unlike scene keypoints, gripper keypoints have known one-to-one correspondences across frames by construction. 
Together, the selected scene and gripper keypoints form correspondence-indexed entities: each selected current-frame keypoint is associated with matched demonstration keypoints under the same correspondence identity. 
These identities serve as the structural prior for our correspondence-conditioned policy.

\vspace{-2mm}
\paragraph{Correspondence-Conditioned Action Prediction.}
Our policy uses the correspondence-indexed entities as an explicit structural prior for action prediction. 
Although the estimated correspondences may still be imperfect, they provide strong cues about which parts of the demonstrations should influence the current scene. 
We organize the input around these correspondence identities. 
Specifically, we group nodes into demonstration frames $\{D_j\}_{j=1}^N$, the current frame $s_c$, and $K$ action frames $(a_k)_{k=1}^K$, where action frames represent future gripper pose and states. 
Each frame contains $M_g=6$ gripper nodes and $M_s=16$ scene nodes. 
Nodes sharing the same index are treated as corresponding entities: the $m$-th scene node in a demonstration frame and the current frame refer to matched scene keypoints, while the $m$-th gripper node across current and action frames refers to the same gripper keypoint. In demonstration and current frames, gripper nodes store 3D gripper-keypoint positions and the gripper state, while scene nodes store selected scene-keypoint positions. 
In action frames, gripper nodes store the predicted displacement of gripper keypoints relative to the current frame, and scene nodes are copied from the current frame as a fixed scene reference.

Since the representation consists of sparse keypoints with structured spatial, temporal, and correspondence relations, a graph structure with message passing provides a natural way to route information among them, as shown on the right of \cref{fig:graph}. 
The information flow is defined by three types of relations. 
First, cross-frame correspondence edges connect each current-frame node to its corresponding nodes in the demonstration frames, and also connect corresponding nodes between adjacent demonstration frames. 
These orange edges in \cref{fig:graph} provide direct pathways for transferring demonstrated intent through matched keypoints. 
Second, intra-frame spatial edges fully connect gripper and scene nodes within each demonstration, current, and action frame. 
These gray edges model the spatial configuration of each frame and allow correspondence information received by scene nodes to propagate to gripper nodes, which are then used for action prediction. 
For these spatial edges, we use the relative Cartesian displacement between nodes as the edge attribute. 
Third, action-frame consistency edges connect gripper nodes in each action frame to their corresponding nodes in the current frame, as well as preceding action frames. 
These blue edges anchor future gripper keypoints to the current observation and encourage temporally consistent predictions. After predicting gripper-keypoint displacements, we recover the relative gripper pose transformation $\Delta T_k$ for each action frame using SVD~\cite{arun1987least}. 
Overall, correspondence information is routed from demonstrations to the current scene, grounded through scene-gripper spatial relations, and decoded into future gripper motion.

\vspace{-1mm}
\paragraph{Local Geometric Refinement.}
The selected keypoints provide only a sparse scene representation, and the estimated correspondences can be noisy or ambiguous. 
To improve robustness, we attach local geometric evidence to each selected scene keypoint. 
Each unselected point in the point cloud is assigned to its nearest selected keypoint, forming local point groups around correspondence-indexed scene nodes. 
These groups are processed by a local PointNet encoder~\cite{qi2017pointnet++}, pre-trained with a geometric reconstruction objective~\cite{mescheder2019occupancy}. 
The resulting features summarize fine-grained local shape evidence around each scene keypoint. 
Through intra-frame spatial edges, these scene-level geometric features are propagated to gripper nodes, allowing local shape evidence to refine action predictions when correspondence information alone is incomplete or unreliable.

\paragraph{Diffusion-based Action Decoding.}
We train the policy with a diffusion-based action decoding objective. 
Given ground-truth gripper-keypoint positions and gripper states in the action frames, Gaussian noise is iteratively added to transform them into a normal distribution. 
The policy predicts the denoising target for each action-frame gripper node:
$\epsilon_\theta=[\nabla p_t,\nabla p_R,\nabla a]\in\mathbb{R}^7$, 
where $\nabla p_t$ and $\nabla p_R$ denote translation- and rotation-induced displacements, and $\nabla a$ denotes the gripper-state residual. 
This decomposition prevents large translational motion from dominating smaller rotational keypoint displacements. 
The model is optimized with the standard denoising MSE objective~\cite{ho2020denoising}.

\subsection{\textsc{MatchingDataset}: Correspondence-Rich In-context Learning Data}
\label{sec:matching_dataset}

\begin{figure}[t]
    \centering
    \includegraphics[width=0.96\columnwidth]{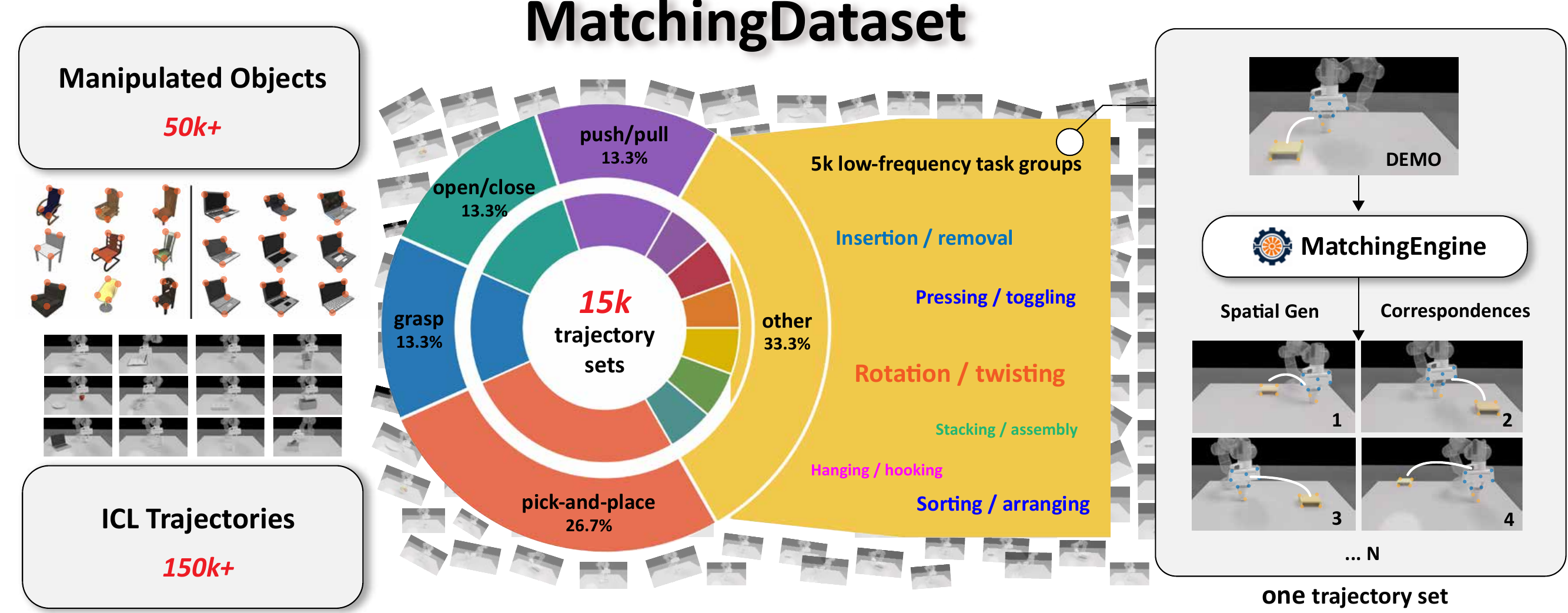}
    \caption{
        Overview of \textsc{MatchingDataset}, 
        \textsc{MatchingEngine} spatially retargets each
        reference demonstration to produce task-consistent trajectories
        with ground-truth 3D correspondences.
    }
    \label{fig:matchingdataset}
\end{figure}

To train an in-context imitation policy, we require intent-consistent trajectories across diverse scene configurations, together with 3D correspondences across the scenes and trajectories. Existing robot datasets, however, are limited in task and object diversity. We therefore design \textsc{MatchingDataset} based on two observations: (i) the policy mainly relies on the relationship between the demonstration and the target scene, rather than object geometry or realistic task appearance; and (ii) cross-scene correspondences are readily available in simulation due to accessible ground-truth poses.
As a result, we build a large-scale synthetic manipulation dataset with 15K intent-consistent task sets, over 150K trajectories, and more than 50K manipulated-object instances. The dataset uses task-relevant meshes sampled from ShapeNet~\cite{chang2015shapenet} and covers both common manipulation behaviors and diverse low-frequency tasks under varied scene configurations, as shown in Fig.~\ref{fig:matchingdataset}.

Each trajectory set is initialized from a procedurally constructed reference demonstration. 
The variant object-centric trajectories are spatially retargeted by adjusting object poses and scene layouts,
while preserving the robot-object relative poses at key frames. As a result, the generated trajectories exhibit diverse observations, robot motions, and spatial configurations, yet maintain the same underlying task intent. Observations are captured as RGB-D point clouds using virtual cameras to mimic realistic manipulation scenarios.
For each object mesh, we pre-sample a dense surface point template in its local coordinate frame, which serves as a canonical representation shared by all trajectories in the same set. Since ground-truth object poses are available in simulation, the template can be transformed into each scene and time step, allowing observed object points to be associated with their corresponding template points. 
Cross-scene and cross-trajectory 3D correspondences are then computed through this shared template space. 
Some template points may not find valid matches in certain time steps in demonstrations within the distance threshold, mimicking real-world partial observations. Each trajectory contains synchronized RGB-D observations, point clouds, and the associated correspondence annotations. Further details on data generation and annotation are provided in Appendix~\ref{app:matchingdataset}.

\subsection{Real-world Correspondence Integration via VLMs}
\label{sec:real}
An advantage of our method is its plug-and-play ability to incorporate off-the-shelf vision foundation models at test time. Given demonstrations and a novel scene of the same task, we sample object-centric candidate points in the initial observations, match them across scenes using VFM features~\cite{heinrich2025radiov2}, propagate the matched points over time with a point tracker~\cite{karaev2024cotracker}, and lift valid 2D matches and tracks to a shared 3D world frame for action prediction.
\begin{figure}[t!]
    \centering
    \includegraphics[width=0.48\textwidth]{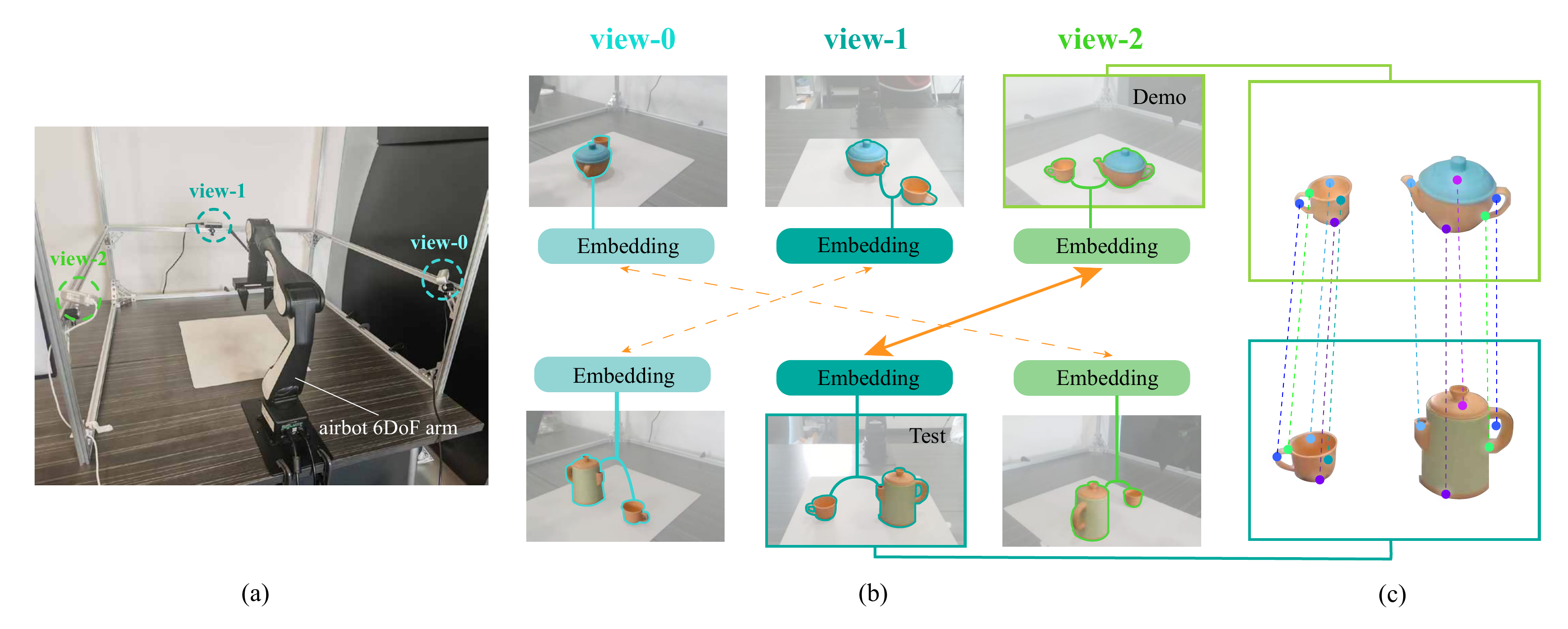} 
    \caption{
    Real-world Setup and Matching
    (a) Camera setup (b) View matching based on the average features of segmented objects.
    (c) Point matching on objects for matched image pairs.}
    \label{fig:matching}\vspace{-1mm}
\end{figure}

\paragraph{Inter-scene Correspondence via Semantic Matching}
\label{inter-scene}

We establish semantic correspondences between the initial frames of the demonstrations and the current scene following the dense feature matching paradigm of recent VFM-based methods~\cite{zhang2023tale}. For each RGB-D observation, we segment the target object region, extract dense descriptors within the object mask, and $L_2$-normalize them before matching. Given a pair of aligned views, we compute 2D correspondences by nearest-neighbor search in descriptor space. For a candidate pixel $u$ in one masked image, its correspondence in the paired image is
\[
v^\ast =
\arg\max_{v \in M_b}
F_a(u)^\top F_b(v),
\]
where $F_a$ and $F_b$ are the normalized dense descriptors of the two images, and $M_b$ is the target object mask in the paired image. The resulting matches provide semantic 2D correspondences between demonstration and current scenes. For multi-object tasks, we apply the same procedure independently within each object mask.

Single-view matching can be unreliable under large viewpoint or object-pose changes due to occlusions and appearance variations. To improve robustness, we perform adaptive view alignment before dense matching. Our setup uses three calibrated RGB-D cameras covering approximately $360^\circ$, providing complementary observations of the scene. Rather than assuming fixed camera-index correspondences, we compare pooled view-level descriptors and select the most compatible view assignment. For a candidate assignment $\pi$, we compute
\[
S(\pi)
=
\frac{1}{|\mathcal{V}|}
\sum_{i \in \mathcal{V}}
\bar{F}_{a,i}^{\top}
\bar{F}_{b,\pi(i)},
\]
where $\bar{F}_{a,i}$ and $\bar{F}_{b,\pi(i)}$ are pooled descriptors from the paired views. We choose $\pi^\ast=\arg\max_{\pi}S(\pi)$ by enumerating feasible assignments, and perform dense matching only on the top two aligned view pairs under $\pi^\ast$. Although our experiments use three cameras, the formulation naturally extends to general multi-view settings.

\paragraph{Intra-scene Correspondence via Visual Tracking}
\label{intra-scene}

Following prior methods~\cite{haldar2025point, wang2025skil}, we use an off-the-shelf point tracker~\cite{karaev2024cotracker} to propagate correspondences over time. Starting from the matched object points in the initial frames, we track their 2D locations through each demonstration sequence offline and through the online execution sequence at test time. Unlike approaches that rely on sparse, expert-annotated, or clustered keypoints, we track a set of object-centric points, with $N_{\text{track}}=512$ points per object and camera view.
Tracking is performed independently for each camera view. For each point, the tracker predicts its 2D location and visibility at each frame, producing temporally consistent image-space trajectories for 3D lifting.

\paragraph{Geometry-based 3D Lifting and Point Filtering}
\label{lifting}
The 3D lifting step is independent of the correspondence source and is applied uniformly to 2D points from semantic matching and visual tracking. We maintain a validity mask for each point: a point is valid only if its depth is available in the RGB-D observation, and tracked points must be predicted as visible by the tracker.
This filtering removes points with missing depth, occlusion, or unreliable tracking. Valid 2D points are then back-projected using depth and camera calibration into the robot-aligned world frame.
Before being passed to the policy, the lifted 3D points at each time step are transformed into the corresponding gripper frame using the inverse gripper pose, yielding temporally consistent 3D point trajectories for downstream manipulation.

\subsection{Implementation Details}
\label{sec:details}
We trained our model using the AdamW optimizer with a learning rate of $1 \times 10^{-5}$.
It took seven days on four NVIDIA A6000 GPUs to train a model for 1M steps. 
All evaluation was performed on a PC with an RTX 4080 GPU.

\section{Experiments}
In this section, we address the following questions: 1) Does \texttt{MatchingPolicy} outperform baseline methods overall as it can better utilize explicit correspondences? 2) Can \texttt{MatchingPolicy} leverage an off-the-shelf large vision model to facilitate generalization in real-world scenarios?



\subsection{Simulation Experiments} 
\paragraph{Evaluation Settings.}
We first evaluate MatchingPolicy on 36 RLBench~\cite{james2020rlbench}
simulation tasks covering diverse objects, initial configurations, and
interaction patterns. We divide these tasks into three groups:
1) $12$ \textbf{Adaptation Tasks}, which are used for fine-tuning to
reduce the domain gap between synthetic pretraining and RLBench;
2) $12$ \textbf{OOD-Easy Tasks}, which are excluded from fine-tuning
but share similar interaction structures with the Adaptation Tasks
(e.g., closing a box versus opening a box); and
3) $12$ \textbf{OOD-Hard Tasks}, which are also excluded from
fine-tuning but involve less familiar interaction patterns, object
affordances, or task semantics.
We compare MatchingPolicy with three representative approaches:
InstantPolicy~\cite{vosylius2024instant}, a geometric in-context learning
method based on graph neural networks;
KAT~\cite{di2024keypoint}, a keypoint-based in-context learning method
that uses LLM reasoning for action generation; and
RDT-1B~\cite{liu2024rdt}, a representative
Vision-Language-Action (VLA) model.
Ground-truth correspondences in RLBench are obtained following the
procedure described in \cref{app:matchingdataset}.

\paragraph{Results.}

\begin{table*}[t!]
\centering
\caption{
Success rates on 36 RLBench tasks.
OOD-Easy (Hard) denotes unseen tasks with similar (dissimilar) interaction patterns to the Adaptation Tasks; 
Results are reported as NoFT/FT, indicating performance without/with fine-tuning on the Adaptation Tasks.
}
\label{tab:comparison}
\setlength{\tabcolsep}{3pt}
\resizebox{1.0\textwidth}{!}{
\begin{tabular}{@{}ccc@{\hspace{0.5cm}}ccc@{\hspace{0.5cm}}ccc@{}}
\toprule
\multicolumn{3}{c}{\textbf{Adaptation Tasks}} &
\multicolumn{3}{c}{\textbf{OOD-Easy Tasks}} &
\multicolumn{3}{c}{\textbf{OOD-Hard Tasks}} \\
\cmidrule(r){1-3} \cmidrule(lr){4-6} \cmidrule(l){7-9}
\textbf{Task} & InstantPolicy & MatchingPolicy &
\textbf{Task} & InstantPolicy & MatchingPolicy &
\textbf{Task} & InstantPolicy\footnotemark[1] & MatchingPolicy \\
\midrule
Open Box           & 0.94/0.99 & 1.00/1.00 & Slide Buzzer       & 0.35/0.94 & 0.71/0.74 & Close Drawer     & --/0.00 & 0.87/0.90 \\
Close Jar          & 0.58/0.93 & 0.78/0.92 & Plate Out          & 0.81/0.97 & 0.96/0.99 & Close Grill      & --/0.00 & 0.98/0.97 \\
Toilet Seat Down   & 0.85/0.93 & 1.00/1.00 & Close Laptop       & 0.91/0.95 & 1.00/1.00 & Open Grill       & --/0.00 & 0.97/0.99 \\
Close Microwave    & 1.00/1.00 & 0.97/1.00 & Close Box          & 0.77/0.99 & 1.00/1.00 & Pick Up Cup      & --/0.00 & 0.90/0.94 \\
Phone on Base      & 0.98/1.00 & 1.00/1.00 & Open Jar           & 0.52/0.78 & 0.89/0.91 & Open Door        & --/0.01 & 0.99/0.98 \\
Lift Lid           & 1.00/1.00 & 1.00/1.00 & Toilet Seat Up     & 0.94/1.00 & 1.00/1.00 & Turn Tap         & --/0.00 & 0.98/1.00 \\
Umbrella Out  & 0.88/0.91 & 0.92/1.00 & Meat off Grill     & 0.77/0.90 & 0.82/0.89 & Frame Off        & --/0.00 & 0.35/0.33 \\
Slide Block        & 0.75/1.00 & 0.98/1.00 & Open Microwave     & 0.23/0.56 & 0.43/0.54 & Sweep Dust       & --/0.00 & 0.37/0.41 \\
Push Button        & 0.60/1.00 & 0.89/1.00 & Paper Roll Off     & 0.70/0.95 & 0.94/0.96 & Open Wine        & --/0.01 & 0.69/0.71 \\
Basketball & 0.66/0.97 & 0.87/0.97 & Put Rubbish & 0.97/0.99 & 1.00/1.00 & Put Money        & --/0.00 & 0.52/0.50 \\
Meat on Grill      & 0.78/1.00 & 0.83/1.00 & Put Umbrella       & 0.31/0.37 & 0.33/0.40 & Put Roll         & --/0.00 & 0.49/0.52 \\
Flip Switch        & 0.40/0.94 & 0.90/0.99 & Lamp On            & 0.42/0.41 & 0.81/0.85 & Pull Plug   & --/0.00 & 0.48/0.46 \\
\midrule
\textbf{Average}   & 0.79/0.97 & 0.92/\textbf{0.99} &
\textbf{Average}   & 0.60/0.82 & 0.83/\textbf{0.86} &
\textbf{Average}   & --/0.00 & 0.72/\textbf{0.73} \\
\bottomrule
\end{tabular}
}
\begin{tablenotes}\small
\item $^1$ We test Instant Policy using the official \href{https://github.com/vv19/instant_policy}{checkpoint}.
\end{tablenotes}
\end{table*}
We first compare MatchingPolicy with InstantPolicy on all 36 RLBench tasks.
As summarized in \cref{tab:comparison}, MatchingPolicy achieves higher average success rates across all three task groups.
A key strength of MatchingPolicy is its ability to generalize without
task-specific fine-tuning. On several unseen tasks, such as
\textit{Open Jar} and \textit{Lamp On}, it even outperforms the baseline
after the latter is fine-tuned on the Adaptation Tasks. This advantage
is most pronounced on OOD-Hard, where the fine-tuned baseline achieves
near-zero performance even when some of the relevant objects have
already appeared during adaptation. We hypothesize that the baseline's
strong reliance on object geometry makes it difficult to infer task
intent when interaction patterns change. In contrast, MatchingPolicy
conditions directly on demonstration-to-scene correspondences, enabling
it to transfer interaction patterns with less dependence on exact object
geometry and generalize more effectively to unseen tasks.

\begin{table}[t]
\centering
\caption{
Comparison with KAT and RDT-1B on six challenging RLBench tasks.
MatchingPolicy achieves the highest average success rate using only two demonstrations.
}
\label{tab:comprehensive_comparison}

\renewcommand{\arraystretch}{1.15}

\begin{threeparttable}
\resizebox{\columnwidth}{!}{%
\begin{tabular}{@{}l|cc|cc|c@{}}
\toprule
\multirow{2}{*}{\textbf{Task}} &
\multicolumn{2}{c|}{\textbf{KAT (ICL)}} &
\multicolumn{2}{c|}{\textbf{RDT-1B (VLA)}} &
\textbf{MatchingPolicy} \\
& $D{=}2$\tnote{1}
& $D{=}10$\tnote{2}
& $D{=}0$\tnote{3}
& $D{=}20$\tnote{2}
& $D{=}2$\tnote{1} \\
\midrule
Plate Out
& 0.12 & 0.36 & 0.08 & 0.67 & \textbf{0.96} \\
Slide Buzzer
& 0.04 & 0.19 & 0.00 & 0.47 & \textbf{0.71} \\
Toilet Seat Up
& 0.25 & 0.38 & 0.12 & 0.69 & \textbf{1.00} \\
Meat off Grill
& 0.33 & 0.54 & 0.05 & 0.73 & \textbf{0.77} \\
Pick Up Cup
& 0.28 & 0.47 & 0.09 & 0.81 & \textbf{0.94} \\
Pull Plug
& 0.00 & 0.12 & 0.00 & 0.32 & \textbf{0.46} \\
\midrule
\textbf{Average}
& 0.17 & 0.34 & 0.06 & 0.62 & \textbf{0.81} \\
\bottomrule
\end{tabular}%
}
\begin{tablenotes}[flushleft]
\scriptsize
\item[] $D$ denotes demo number.
\textsuperscript{1} Our setup;
\textsuperscript{2} original method setup;
\textsuperscript{3} zero-shot setup.
\end{tablenotes}
\end{threeparttable}\vspace{-1.5em}
\end{table}
We further compare MatchingPolicy with two representative baselines on
six challenging tasks: KAT, an LLM-based in-context learning
method~\cite{di2024keypoint}, and RDT-1B, a representative VLA
model~\cite{liu2024rdt} (\cref{tab:comprehensive_comparison}).
Using only two demonstrations, MatchingPolicy consistently outperforms
KAT, even when KAT is provided with ten demonstrations.
It also achieves higher success rates than RDT-1B, even after the latter
is fine-tuned on 20 demonstrations.
Together, these results highlight the effectiveness and demonstration
efficiency of explicit correspondence modeling compared with LLM-based
in-context learning and VLA fine-tuning.

\subsection{Real-World Experiments}
\paragraph{Setup and evaluation Settings.}
We evaluate MatchingPolicy on four real-world manipulation tasks spanning rigid-object and articulated-object manipulation. The experiments are conducted using an AirBot Pro 6-DoF robotic
arm\footnote{\url{https://airbots.online/}}, with the scene captured by three Intel RealSense D435i RGB-D cameras.
MatchingPolicy extracts visual and semantic correspondences using
off-the-shelf vision foundation models and is deployed directly without real-world fine-tuning.
To rigorously evaluate generalization, we consider three levels of difficulty:
\textbf{Layout-Easy}, which uses the same object instances as the
demonstrations with minor pose variations;
\textbf{Layout-Hard}, which uses the same instances with substantial
pose variations, including markedly different orientations; and
\textbf{Shape}, which uses novel object instances that differ in
appearance and geometry.
For each task and evaluation setting, we conduct ten trials and report
the average success rate.
Detailed task illustrations and success criteria are provided in
\cref{sec:real_details}.
\begin{figure*}[t!]
    \centering
    \includegraphics[width=0.82\textwidth]{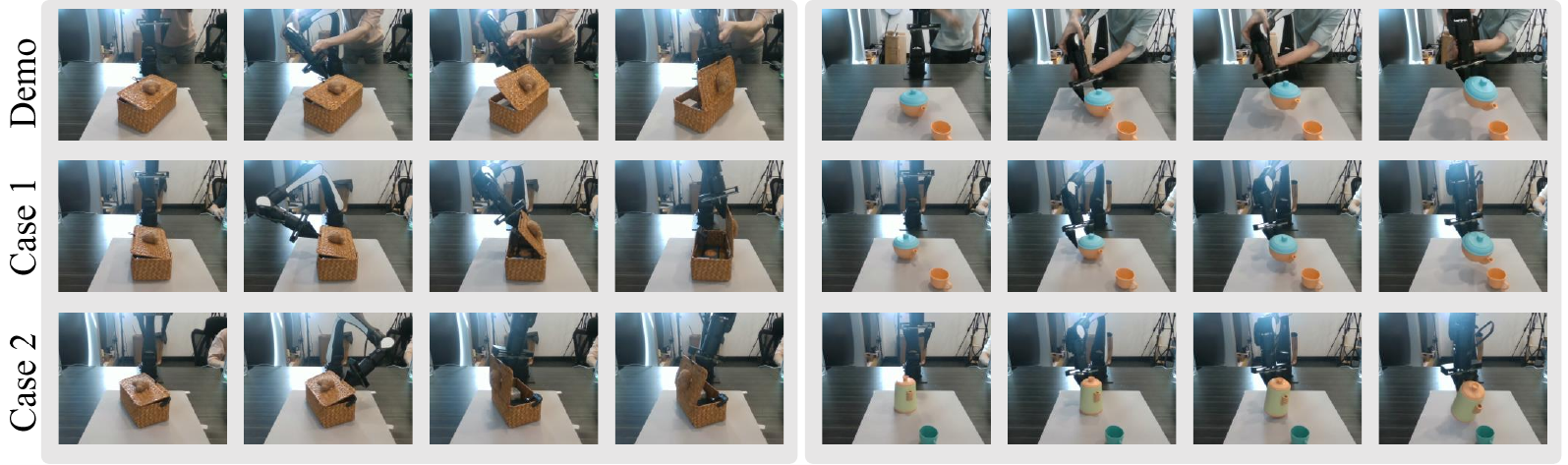} 
    \caption{Generalization capabilities of MatchingPolicy with a single demonstration. Left: Novel layouts; Right: Unseen object instances} 
    \label{fig:demo}
\end{figure*}

\begin{table}[t!]
\centering
\caption{Results on real-world tasks, evaluated across different levels of generalization.}
\label{tab:real}

\fontsize{8.5pt}{9.5pt}\selectfont
\setlength{\tabcolsep}{3.8pt}
\renewcommand{\arraystretch}{1.15}

\begin{tabular}{@{}llccccc@{}}
\toprule
\textbf{Level} & \textbf{Method}
& \shortstack{\textbf{Pour}\\\textbf{Water}}
& \shortstack{\textbf{Put}\\\textbf{Lid}}
& \shortstack{\textbf{Open}\\\textbf{Box}}
& \shortstack{\textbf{Cut}\\\textbf{Egg}}
& \textbf{Averge} \\
\midrule

\multirow{2}{*}{Layout-Easy}
& InstantPolicy
& 0/10
& \textbf{9/10}
& 0/10
& 0/10
& 22.5\% \\
& MatchingPolicy
& \textbf{7/10}
& 8/10
& \textbf{9/10}
& \textbf{7/10}
& \textbf{77.5\%} \\

\midrule
\multirow{2}{*}{Layout-Hard}
& InstantPolicy
& 0/10
& 0/10
& 0/10
& 0/10
& 0.0\% \\
& MatchingPolicy
& \textbf{6/10}
& \textbf{6/10}
& \textbf{8/10}
& \textbf{6/10}
& \textbf{65.0\%} \\

\midrule
\multirow{2}{*}{Shape}
& InstantPolicy
& 0/10
& 4/10
& 0/10
& 0/10
& 10.0\% \\
& MatchingPolicy
& \textbf{8/10}
& \textbf{6/10}
& \textbf{8/10}
& \textbf{7/10}
& \textbf{72.5\%} \\

\bottomrule
\end{tabular}
\vspace{-1.5em}
\end{table}
\paragraph{Overall}
As shown in \cref{tab:real}, MatchingPolicy substantially outperforms
InstantPolicy across all three evaluation settings.
Its performance remains robust under both layout and object variations,
demonstrating strong real-world generalization without real-world
fine-tuning.

\paragraph{Generalization to novel layout and novel objects}
We observe that the success rate of InstantPolicy on the \textit{Put Lid} task in the \textbf{Layout-Hard} setting drops to $0\%$. Relying on a coarse geometric representation and biased by the demonstration trajectory, it fails by grasping the pot instead of the lid. In contrast, our approach leverages VFMs to extract precise semantic correspondences, enabling MatchingPolicy to identify the lid and complete the task.
In more challenging task \textit{Open Box}, MatchingPolicy can generate successful trajectories approaching objects from directions totally different from that in the demonstration (\cref{fig:demo}). This shows our method follows the motion of matched keypoints instead of merely replaying the demonstrated trajectory.

In the \textbf{Shape} generalization setting, MatchingPolicy again significantly outperforms the baseline, achieving a $72.5\%$ average success rate. As illustrated on the right of \cref{fig:demo}, for example, MatchingPolicy successfully completes the \textit{Pour Water} task even when the teapot differs substantially in shape. 
Remarkably, the policy also demonstrates cross-category generalization on the \textit{Put Lid} task, correctly placing a lid on a teapot after seeing only a demonstration of placing a lid onto a pot (see our supplemental materials). These results confirm that our method relies on learned correspondences and is largely agnostic to object geometry.

\subsection{Ablation Studies}
We conduct comprehensive ablation studies by comparing several variants of our approach.
We evaluate our method against variants using a naive matching approach in real-world scenarios. Additionally, we explore test-time planning strategies to further enhance our policy's performance.

\paragraph{Two-Stage Matching versus Naive 2D-to-3D Matching}
We compare our two-stage matching approach against a 3D naive matching baseline. The baseline, based on \cite{wang2024gendp}, projects 2D features into the world frame and fuses them by their 3D coordinates before establishing correspondences. As shown in Tab.~\ref{tab:ablation}, our method achieves superior performance by generating more accurate semantic correspondences. We observe that the performance of the naive baseline drops in the \textbf{Shape} and \textbf{Layout-Hard} settings, as naive matching produces many incorrect correspondences which mislead the policy and lead to task failure.

\paragraph{Test-time Planning Methods} 
We develop a planning strategy adapted from \cite{janner2022diffuser} to improve test-time performance: 
After the policy predicts $K$ actions, the robot executes the first $K/2$ actions.
The remaining $K/2$ actions seed the next prediction by replacing the first $K/2$ action of the next action chunk, resulting in an inference process where the first half of an action chunk is denoised for only the final step, while the second half undergoes the full diffusion process.
\begin{table}[t]
\centering
\footnotesize
\setlength{\tabcolsep}{3pt}
\caption{Average success rates of our  model and ablated variants.}
\resizebox{0.9\columnwidth}{!}{
\begin{tabular}{@{}ccccc@{}}
\toprule
\textbf{Matching} & \textbf{Planning} & \textbf{Layout-Easy} & \textbf{Layout-Hard} & \textbf{Shape} \\
\midrule
Naive & Overlap & 60.0\% & 20.0\% & 45.0\% \\
2-Stage & Whole & 54.0\% & 47.5\% & 57.5\% \\
2-Stage & Half & 60.0\% & 45.0\% & 45.0\% \\
2-Stage & Overlap & 77.5\% & 65.0\% & 72.5\% \\
\bottomrule
\end{tabular}
}
\label{tab:ablation}
\end{table}
We term this strategy ``overlapping chunks'' and use it in other real-world experiments.
We compare it against several alternatives: 1) Whole chunk (``Whole''): the robot executes all $K$ predicted actions. 2) Half chunk (``Half''): the robot executes the first $K/2$ steps and then re-predicts.
The results in \cref{tab:ablation} demonstrate that ``overlap'' significantly enhances policy performance.
In addition, it improves the smoothness of the resulting trajectories.

\section{Conclusions and Limitations}

\paragraph{Conclusions} In this work, we introduced MatchingPolicy, a novel correspondence-aware policy that transforms robot actions from demonstrations to new scenes using dense correspondence cues. By leveraging off-the-shelf vision foundation models with a novel two-stage matching algorithm, our approach enables one-shot imitation on tasks with unseen objects and different layouts in the real world. Our work demonstrates that explicitly modeling the visual and semantic relationships between the demonstration and the current scene is key to effective in-context imitation learning. This principle of learning from analogy, rather than from vast datasets of robot trials, paves the way for building more generalizable and data-efficient robotic systems.

\paragraph{Limitations} Our method also has limitations.
First, we focus primarily on short-horizon manipulation tasks; extending our approach to long-horizon tasks would be a valuable direction for future work.
Second, our method considers semi-dynamic scenes and operates at a relatively slow speed, making it difficult to handle highly dynamic tasks such as fling~\cite{ha2022flingbot}.
Third, our view selection process is somehow dependent on our current platform,
requiring adaptation on other platforms. We will address them in future work.
{
    \small
    \bibliographystyle{ieeenat_fullname}
    \bibliography{main}
}

\clearpage
\section{\textsc{MatchingDataset} Generation Details}
\label{app:matchingdataset}
\begin{figure*}[t!]
    \centering
    \includegraphics[width=0.9\textwidth]{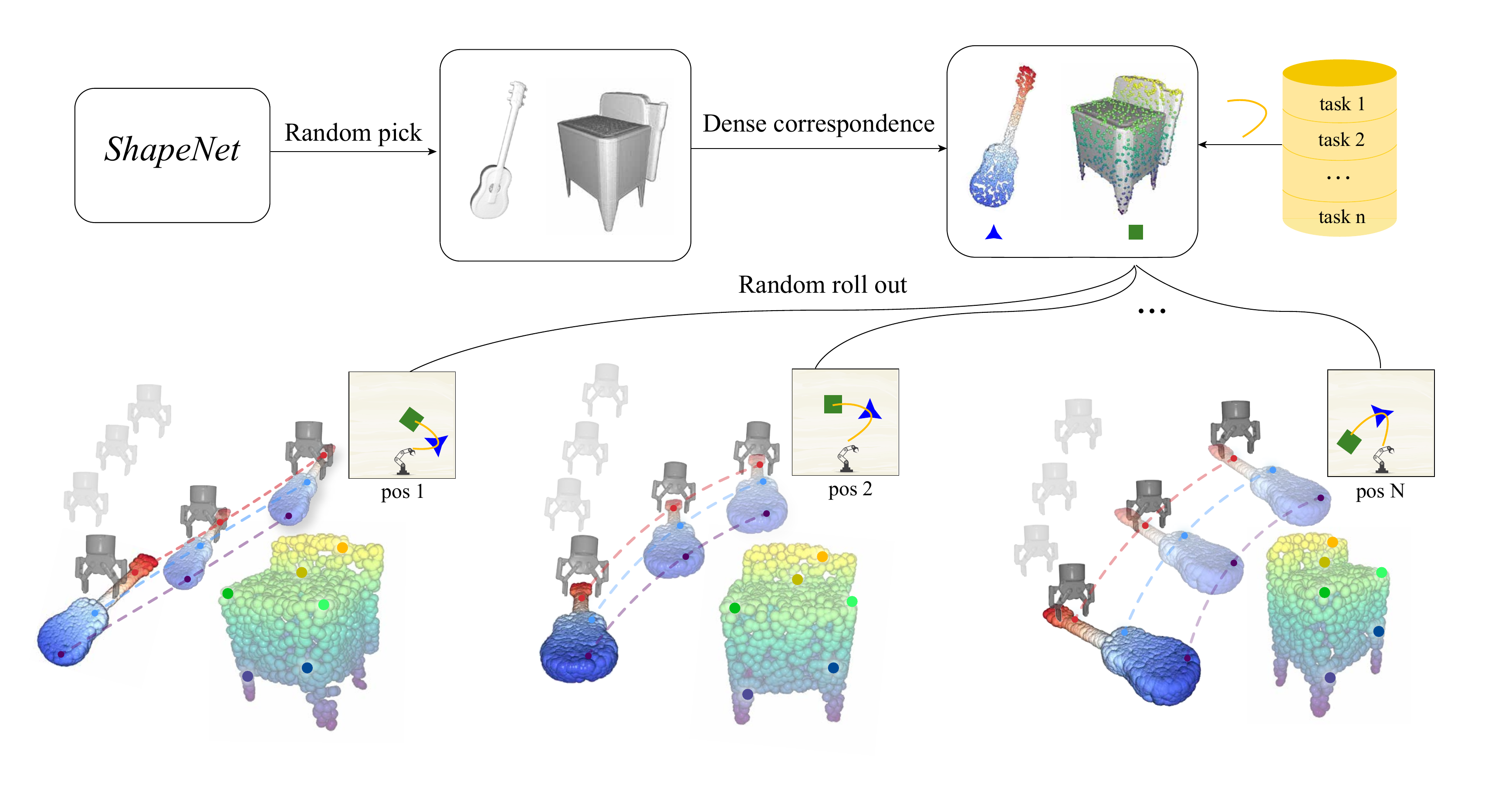} 
    \caption{Data generation pipeline of MatchingDataset} 
    \label{fig:sythetic}
\end{figure*}
\subsection{Dataset Organization}

\textsc{MatchingDataset} contains 15K task-consistent trajectory sets
and over 150K trajectory realizations, involving more than 50K
manipulated-object instances. Each set is initialized from one
procedurally constructed reference demonstration. We organize the
dataset as
\begin{equation}
    \mathcal{D}
    =
    \left\{
        \mathcal{S}_m
    \right\}_{m=1}^{M},
    \qquad
    \mathcal{S}_m
    =
    \left\{
        \tau_m^{(n)}
    \right\}_{n=1}^{N_m},
    \label{eq:dataset_organization}
\end{equation}
where $\mathcal{D}$ denotes the complete dataset, $M$ is the number of
task-consistent sets, and $m$ indexes these sets.
$\mathcal{S}_m$ is the set initialized from the $m$-th reference
demonstration, $N_m$ is its number of trajectory realizations, and
$\tau_m^{(n)}$ denotes its $n$-th trajectory.

Trajectories within $\mathcal{S}_m$ preserve the same task specification
and object models while varying object poses, scene layouts,
observations, and robot motions. Each trajectory contains synchronized
RGB-D observations, point clouds, robot and gripper states, and
end-effector actions, together with ground-truth 3D correspondence and
visibility annotations. 

\subsection{Reference Synthesis and Spatial Retargeting}

As shown in~\cref{fig:sythetic}, we construct reference demonstrations using task-relevant object meshes
sampled from ShapeNet~\cite{chang2015shapenet}. Each demonstration is
defined by an object-centric task specification describing the roles of
the involved objects, their interaction regions, and the desired spatial
relation or motion. These specifications instantiate common behaviors
such as grasping, pick-and-place, pushing or pulling, and opening or
closing, as well as lower-frequency behaviors including insertion,
rotation, pressing, stacking, hanging, and sorting.

We convert each task specification into object-centric end-effector
waypoints and corresponding gripper states, which are interpolated to
produce a continuous trajectory. For prehensile interactions, the
manipulated object maintains a fixed relative transform to the gripper
after grasping. The resulting motion is synthesized kinematically and
rendered from the configured camera views to obtain synchronized
observations and actions.

Given a reference demonstration, \textsc{MatchingEngine} generates
additional realizations through object-centric spatial retargeting.
Similar object-centric retargeting principles are used in
MimicGen~\cite{mandlekar2023mimicgen} and
DemoGen~\cite{xue2025demogen}. Specifically, we sample new poses for
task-relevant objects and perturb the surrounding scene layout. Because
each waypoint is expressed in the coordinate frame of its associated
object, it can be transformed together with the object while preserving
the corresponding interaction region.

For multi-object tasks, different trajectory segments may be anchored to
different objects. For example, the grasp segment of a pick-and-place
trajectory is anchored to the source object, whereas the placement
segment is anchored to the target object. We transform each segment
using its associated object pose and connect the resulting segments
through interpolated free-space motion. The reference demonstration and
its retargeted realizations jointly form one task-consistent set
$\mathcal{S}_m$.

\subsection{Ground-Truth Correspondence Annotation}

All trajectories within a task-consistent set retain the same object
meshes and local object coordinates. This shared object space allows
canonical surface points to define correspondence identities across
frames and trajectory realizations.

For clarity, we consider one task-relevant object and omit the set and
object indices. We uniformly sample $W=2048$ canonical surface anchors
$\{\mathbf{x}_j\}_{j=1}^{W}$ from its mesh. The world-space position of
anchor $\mathbf{x}_j$ in frame $t$ of trajectory $n$ is
\begin{equation}
    \mathbf{p}_{j,t}^{(n)}
    =
    \mathbf{R}_{t}^{(n)}\mathbf{x}_j
    +
    \mathbf{t}_{t}^{(n)},
    \label{eq:gt_correspondence}
\end{equation}
where $\mathbf{R}_{t}^{(n)}$ and $\mathbf{t}_{t}^{(n)}$ denote the
frame-wise object rotation and translation, respectively. Points
transformed from the same canonical anchor $\mathbf{x}_j$ therefore
share an exact correspondence identity across frames and trajectories.

Let $\mathcal{P}_{t}^{(n)}$ denote the observed point cloud in the same
frame. We associate each transformed anchor with its nearest observed
point:
\begin{equation}
    \mathbf{q}_{j,t}^{(n)}
    =
    \underset{\mathbf{q}\in\mathcal{P}_{t}^{(n)}}{\arg\min}
    \left\|
        \mathbf{q}-\mathbf{p}_{j,t}^{(n)}
    \right\|_2.
    \label{eq:nearest_observed_point}
\end{equation}
We mark the anchor as visible when
\begin{equation}
    \left\|
        \mathbf{q}_{j,t}^{(n)}
        -
        \mathbf{p}_{j,t}^{(n)}
    \right\|_2
    <
    r_{\mathrm{vis}},
    \label{eq:visibility}
\end{equation}
where $r_{\mathrm{vis}}$ is the visibility tolerance. Because the point
cloud is back-projected from the rendered depth observation, occluded or
out-of-view anchors do not have a sufficiently close observed point.
For every frame, we store the canonical anchor identity, its associated
observed point, and its visibility flag. This provides ground-truth
cross-trajectory correspondence supervision without requiring a learned
feature matcher during training.

\section{Real-world Experiment Details and Results}
\label{sec:real_details}
\begin{figure}[t!]
    \centering
    \includegraphics[width=0.96\columnwidth]{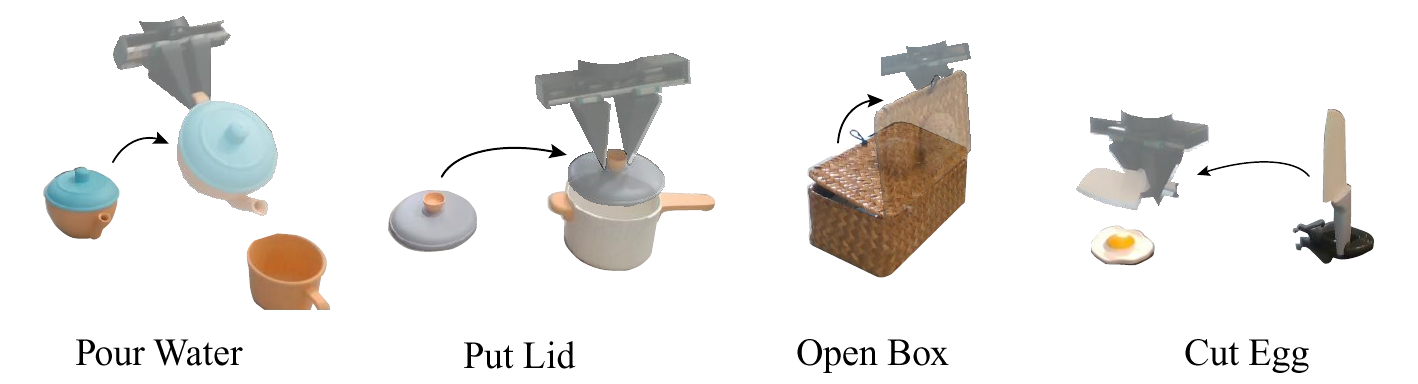} 
    \caption{The real-world four tasks for evaluation.}
    \label{fig:tasks}
\end{figure}

\paragraph{Tasks and Success Criteria} We design four real-world tasks (~\cref{fig:tasks}) and define \textbf{strict} success criteria for each task: \textbf{Pour Water:} the teapot spout is positioned vertically above the cup and the teapot has been tilted by ${\geq 15^\circ}$; \textbf{Put Lid:} the lid covers $ \geq 60\%$ of the pot's opening; \textbf{Open Box:} the lid has been completely flipped open; \textbf{Cut Egg:} The knife has reached the top of the egg yolk and has then moved downward by at least $2cm$.

\begin{figure}[t!]
    \centering
    \includegraphics[width=0.96\columnwidth]{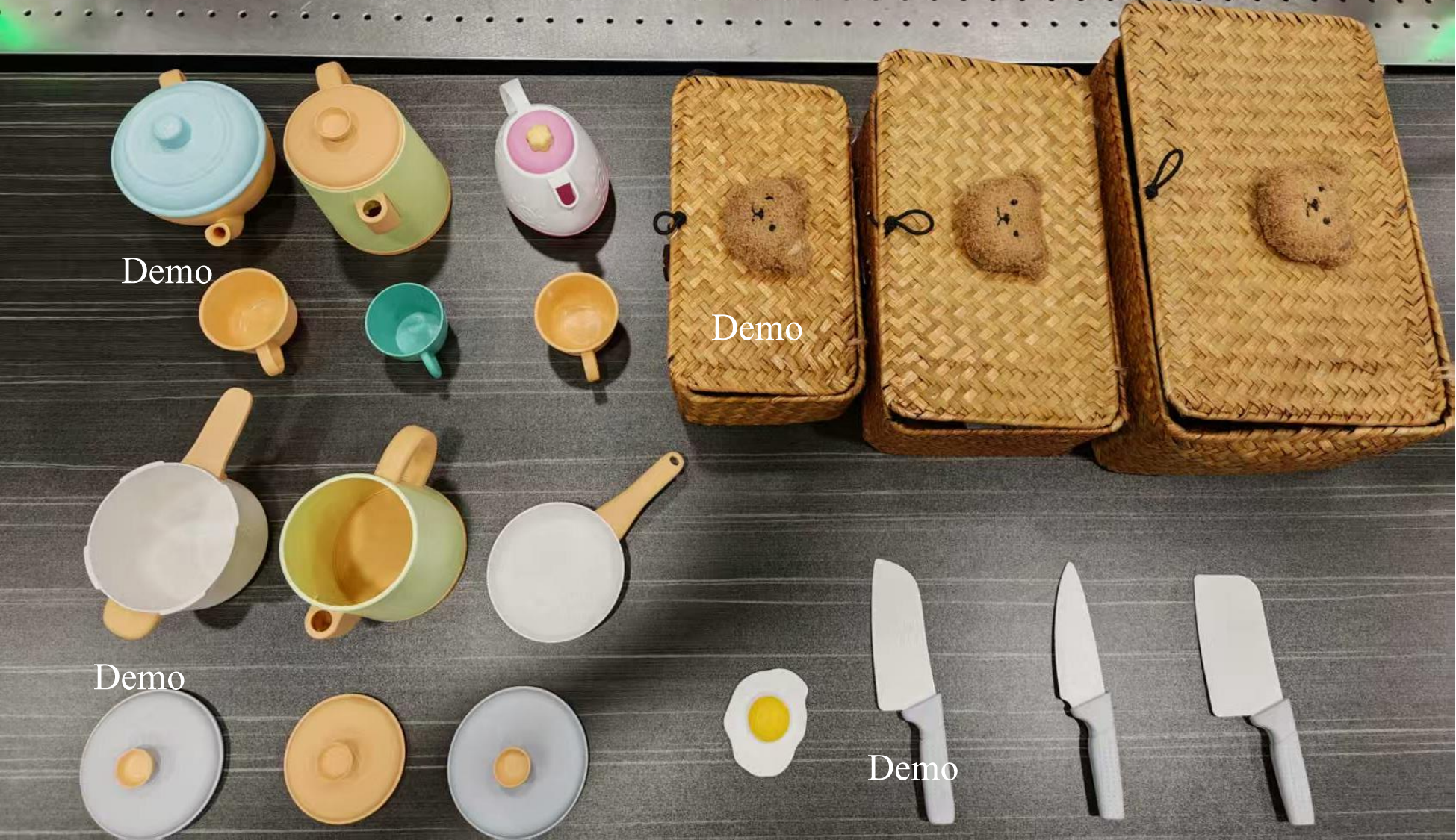} 
    \caption{The diverse set of test objects}
    \label{app:var} 
\end{figure}

\paragraph{Three-level Generalization}
We assess our models on a set of challenging everyday manipulation tasks, as shown in \cref{fig:tasks}. The object variants are shown in \cref{app:var}.
\textbf{Layout-Easy:} We randomize the scene by rotating all objects between $-30^\circ$ and $30^\circ$ and applying a random translational offset of 3--5 cm on the table plane; \textbf{Layout-Hard:} We apply more aggressive randomization, rotating objects between $90^\circ$ and $150^\circ$ with the same 3--5 cm translational offset; \textbf{Shape:} We replace the demonstration objects with semantically similar counterparts that differ in appearance and geometry (~\cref{app:var}), and then randomize the scene according to the ``Layout-Easy" setting.
\begin{figure*}
    \centering
    \includegraphics[width=0.96\textwidth]{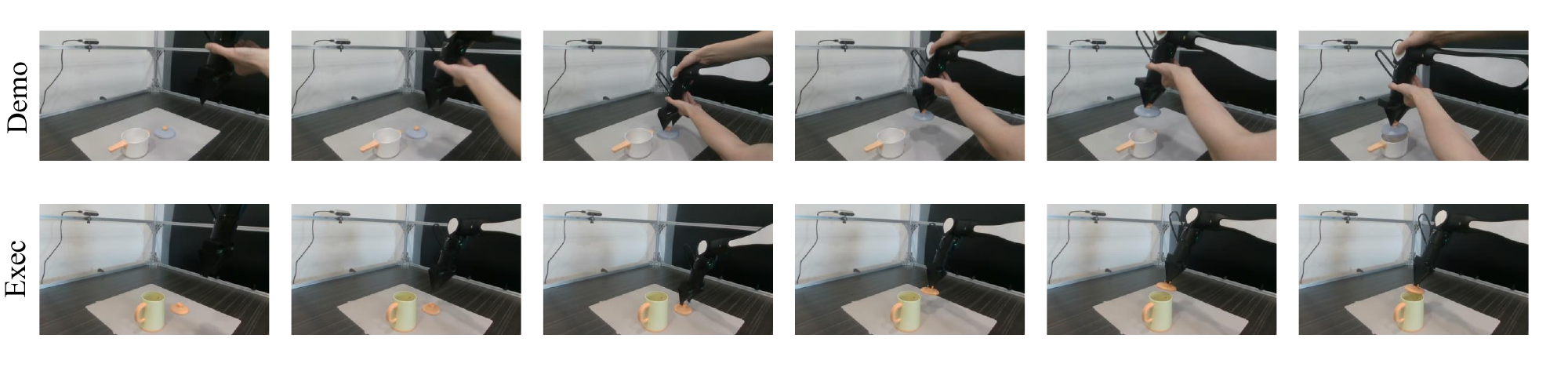} 
    \caption{Cross-Category Generalization}
    \label{app:cross} 
\end{figure*}
\paragraph{Inter-Category Generalization} For example, given a demonstration of placing a lid onto a pot, our policy can successfully adapt to place a lid onto a novel teapot (See ~\cref{app:cross}).

\begin{figure*}
    \centering
    \includegraphics[width=0.92\textwidth]{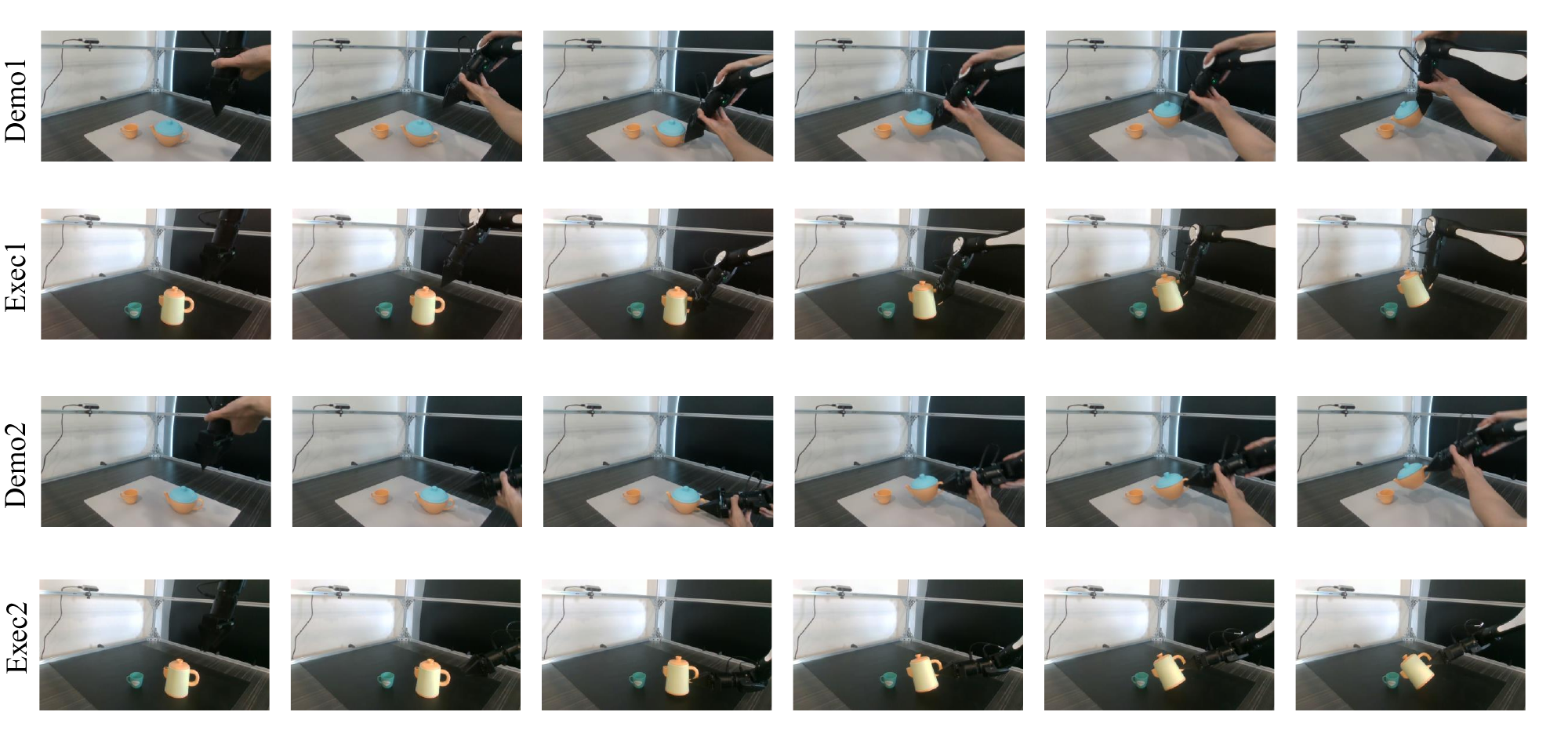} 
    \caption{Interaction following shown in the demonstrations.}
    \label{app:follow} 
\end{figure*}
\paragraph{Interaction Following} To test the nuanced understanding of our policy, we provided two demonstrations of the \textbf{Pour water} task that \textit{differed only in the grasping style}. Our model not only succeeded at the task but also precisely replicated the distinct interaction style from each demonstration (See ~\cref{app:follow}).
\end{document}